\documentclass[10pt,twocolumn,letterpaper]{article}

\usepackage{cvpr}

\usepackage{multirow}
\usepackage[table]{xcolor}

\usepackage{booktabs}
\usepackage{colortbl}

\definecolor{secblue}{RGB}{230,241,251}
\definecolor{secgreen}{RGB}{234,243,222}
\definecolor{bestrow}{RGB}{255,243,205}

\definecolor{cvprblue}{rgb}{0.21,0.49,0.74}
\usepackage[pagebackref,breaklinks,colorlinks,allcolors=cvprblue]{hyperref}

\def\paperID{*****} 
\def\confName{CVPR}
\def\confYear{2026}

\title{Towards Fair and Robust Volumetric CT Classification via KL-Regularised Group Distributionally Robust Optimisation}

\author{
Samuel Johnny$^{1}$\thanks{Both authors contributed equally to this research.},
Blessed Guda$^{2}$\footnotemark[1],
Goodness Obasi$^{1}$,
Aaron Emmanuel$^{1}$,
Moise Busogi$^{1}$\\
$^{1}$Carnegie Mellon University Africa, Kigali, Rwanda\\
$^{2}$Carnegie Mellon University, Pittsburgh, USA\\
{\tt\small \{sjohnny, blessedg, gobasi, eaaron, mbusogi\}@andrew.cmu.edu}
}

\begin{document}
\maketitle
\begin{abstract}
Automated diagnosis from chest computed tomography (CT) scans faces two persistent challenges in clinical deployment: distribution shift across acquisition centres, and performance disparity across demographic subgroups. We address both simultaneously across two complementary tasks: binary COVID-19 classification from multi-site CT volumes (\textit{Task 1}) and four-class lung pathology recognition with gender-based fairness constraints (\textit{Task 2}). Our framework combines a lightweight MobileViT-XXS slice encoder with a two-layer SliceTransformer aggregator for volumetric reasoning, and trains with a KL-regularised Group Distributionally Robust Optimisation (GDRO) objective that adaptively upweights underperforming acquisition centres and demographic subgroups. Unlike standard GDRO, the KL penalty prevents group weight collapse, providing a stable balance between worst-case protection and average performance. For \textit{Task 2}, we define groups at the granularity of gender$\times$class, directly targeting severely underrepresented combinations such as female Squamous cell carcinoma. On \textit{Task 1}, our best configuration achieves a challenge F1 of \textbf{0.835}, surpassing the best published challenge entry by $+5.9$. On Task 2, GDRO with $\alpha = 0.5$ achieves a mean per-gender macro F1 of \textbf{0.815}, outperforming the best challenge entry by $+11.1$ and improving Female Squamous F1 by $+17.4$ over the Focal Loss baseline.

\end{abstract}    
\section{Introduction}
\label{sec:intro}

Early detection of diseases such as COVID-19 and primary lung carcinomas from computed tomography (CT) scans has attracted significant research attention, owing to the potential for scalable, automated clinical screening~\cite{kollias2022ai, kollias2021mia, kollias2023ai}. While early approaches treated each CT volume as a collection of independent 2D axial slices, this discards the inter-slice spatial context that is often critical for diagnosis~\cite{kollias2023deep}. More recent methods process the full 3D volume directly, though this incurs substantial computational cost. A practical middle ground is to encode slices independently with a shared 2D backbone and then aggregate the resulting representations into a single volumetric descriptor, preserving spatial context at manageable cost.

Despite their promise, CT-based classifiers face two persistent challenges when deployed in real clinical settings. First, \textit{domain shift}: scans acquired across different hospitals vary in scanner hardware, reconstruction kernels, and imaging protocols, causing models trained on one site to generalise poorly to others~\cite{gerogiannis2024covid}. Second, \textit{demographic fairness}: class distributions are rarely balanced across patient subgroups such as gender, and standard training objectives optimise average performance at the expense of underrepresented groups. These issues are especially pronounced in small, multi-centre datasets, where existing domain generalisation methods, typically designed for large-scale settings, offer limited
benefit~\cite{guan2022domain,zhang2020generalizing}.

In this work, we address both challenges simultaneously through a group-aware training objective applied to two related tasks: binary COVID-19 classification across four acquisition centres (\textit{Task~1}), and four-class lung pathology classification with gender-based fairness constraints (\textit{Task~2})~\cite{kollias2025pharos, kollias2023ai}. We build on Group Distributionally Robust Optimisation (Group DRO)~\cite{sagawa2020distributionally}, which minimises the worst-case loss over predefined subgroups rather than the average loss, and augment it with a KL regularisation term that prevents the group weights from collapsing onto a single hard group.
This KL-regularised GDRO objective provides a principled mechanism for reweighting underrepresented subgroups such as female Squamous cell carcinoma patients (5 training samples), without discarding information from majority groups.

Our framework is built on a lightweight MobileViT-XXS slice
encoder~\cite{mehta2022mobilevit} paired with a two-layer SliceTransformer aggregator, keeping the total parameter count under 1.3M and making the approach practical for small clinical datasets.
\noindent In summary, our contributions are:
\begin{enumerate}
  \item A volumetric CT classification framework combining a
        MobileViT-XXS slice encoder with a SliceTransformer
        aggregator, achieving competitive performance at under
        1/3M parameters on both single-site and multi-centre
        benchmarks.

  \item A KL-regularised GDRO training objective that adaptively reweights acquisition centres (\textit{Task~1}) and fine-grained gender$\times$class subgroups (\textit{Task~2}), improving worst-case subgroup performance over standard weighted cross-entropy and Focal Loss baselines~\cite{lin2017focal}.
  \item Systematic ablation of the KL regularisation strength $\alpha$ and aggregator design, providing actionable insights for practitioners working with small, demographically imbalanced medical imaging datasets.
\end{enumerate}
\section{Related Work}
\label{sec:related}
 
\subsection{CT-based Disease Classification}
 
Automated analysis of chest CT scans for disease detection has
been an active area of research, driven by the clinical need
for scalable screening tools.
Early approaches processed CT volumes as independent collections
of 2D axial slices, applying standard convolutional neural
networks to each slice before aggregating predictions via
majority voting or simple
pooling~\cite{arsenos2022large, kollias2018deep,ALMANSOR2023234,wang2023enhancing}.
While computationally efficient, these methods discard
inter-slice spatial context that is often critical for
distinguishing pathologies with characteristic 3D
distributions, such as ground-glass opacities in
COVID-19~\cite{kollias2021mia,10.1145/3644116.3644135}.
 
The emergence of large-scale CT datasets has enabled more
sophisticated volumetric approaches.
Kollias et al.~\cite{kollias2022ai} introduced the AI-MIA
framework for COVID-19 detection and severity grading,
demonstrating that 3D-aware architectures outperform
slice-level baselines when sufficient training data is
available.
Subsequent work extended this to multi-source settings,
combining CT with clinical metadata and achieving improved
generalisation across
multipele data centres~\cite{kollias2023ai, gerogiannis2024covid}.
More recently, vision-language models have been explored for
CT segmentation and detection, leveraging large pretrained
representations to reduce the annotation burden~\cite{kollias2024sam2clip2sam}.
Our work builds on this line of research but focuses on the under-explored regime of small, multi-centre datasets where volumetric encoders must be both lightweight and
distribution-aware.
 
\subsection{Domain Generalisation in Medical Imaging}
 
Distribution shift across data centers is one of the
most persistent challenges in clinical AI deployment.
Scanners from different manufacturers, varying reconstruction
kernels, and institutional imaging protocols all contribute
to covariate shift that degrades model performance when
evaluated on unseen
domains~\cite{guan2022domain, kollias2025pharos,10780969}.
Guan and Liu~\cite{guan2022domain} provide a comprehensive
survey of domain adaptation techniques for medical imaging,
cataloguing approaches ranging from adversarial feature
alignment to self-supervised pretraining. The survey showed
that most methods require either large multi-site training sets or access to target domain data at test time.
 
Data augmentation strategies have been proposed to simulate
domain diversity during training. Zhang et al.~\cite{zhang2020generalizing} demonstrate that stacking domain-randomised transformations can improve generalisation to unseen sites for segmentation tasks, but find that performance degrades significantly when the training set is small and lacks site diversity~\cite{zhang2020generalizing, ZHANG2026103848}.
This limitation is particularly acute in our setting, where
the training data comprises fewer than 1{,}300 volumes across
four centres with highly unequal class distributions. Rather than augmenting the input space, we address distribution shift directly in the objective function via group-level loss reweighting, which does not require additional data or
target domain samples.
 
\subsection{Distributionally Robust Optimisation}
 
Distributionally Robust Optimisation (DRO) provides a
principled framework for training models that perform well
under worst-case distributional
shifts~\cite{sagawa2020distributionally,esfahani2018wasserstein,sagawa2020overparameterization}.
Group DRO, introduced by Sagawa et al.~\cite{sagawa2020distributionally},
minimises the maximum expected loss over a set of predefined
groups, rather than the average loss across all samples.
Group weights are updated by exponentiated gradient ascent,
upweighting groups with high loss at each training step.
Empirically, Group DRO has been shown to substantially reduce
worst-case subgroup error on benchmark datasets, though the
authors note that strong $\ell_2$ regularisation is necessary
to prevent overfitting on small minority groups.
 
Follow-up work has applied DRO to long-tail recognition,
fairness in face recognition and medical imaging
fairness~\cite{kollias2025pharos}.
Kollias et al.~\cite{kollias2025pharos} specifically address
multi-source and demographic fairness in disease diagnosis,
demonstrating that group-aware objectives improve per-subgroup
performance on CT-based tasks.
Our work extends this line by introducing a KL regularisation
term toward the uniform group weight distribution, which
prevents the group weights from collapsing onto a single
hard group, a failure mode observed at high DRO learning
rates on small datasets.
Furthermore, we define groups at the finer granularity of
gender$\times$class for \textit{Task~2}, directly targeting
underrepresented combinations such as female Squamous cell
carcinoma (5 training samples) that would otherwise be
subsumed by majority group gradients.
 
\subsection{Lightweight Vision Transformers for Medical Imaging}
 
The Vision Transformer (ViT)~\cite{dosovitskiy2021image} and its
descendants have demonstrated strong performance on the image
classification benchmarks, but their quadratic attention
complexity, large parameter count, and large data size make them impractical for resource-constrained medical imaging pipelines.
MobileViT~\cite{mehta2022mobilevit} addresses this by
interleaving standard MobileNet convolution blocks with local
Transformer blocks, achieving competitive accuracy at a
fraction of the parameters of standard ViT models.
The \textit{XXS} variant used in our work contains approximately
1.3M parameters in the backbone, making it well-suited for
small dataset regimes where large models are prone to
overfitting.
 
Several recent works have adapted lightweight transformers
for medical imaging~\cite{huang2026inceptionmamba,zhang2023swipe,11149776,lee2025multisource,yuan2025multisourcevre,li2025advancing}.
Kollias et al.~\cite{kollias2023deep} propose a deep neural architecture that harmonises 3D input analysis with decision making, demonstrating that compact architectures with task-specific aggregation outperform heavier general-purpose models when training data is
limited~\cite{kollias2020deep, kollias2020transparent}. Similarly, SAM2CLIP2SAM~\cite{kollias2024sam2clip2sam} proposes
a vision-language pipeline for 3D CT segmentation in which a
Segment Anything Model (SAM) first delineates candidate lung
regions, CLIP provides semantic grounding of the segmented
regions, and a second SAM pass refines the final segmentation
masks for COVID-19 detection. While this approach leverages powerful pretrained vision-language representations to reduce annotation
dependence, it is designed for segmentation, which introduces an overhead that may not be needed for this data. It also does not address distribution shift across different groups.
Our SliceTransformer aggregator follows this philosophy:
rather than processing the full 3D volume with a volumetric
backbone, we apply a shared 2D encoder to each slice and
use a lightweight two-layer Transformer Encoder to aggregate
slice-level features.
This design keeps the total parameter count low while
capturing long-range inter-slice dependencies that are simple
pooling aggregators miss.

\begin{figure}[htbp]
\centering
\includegraphics[width=1.0\linewidth]{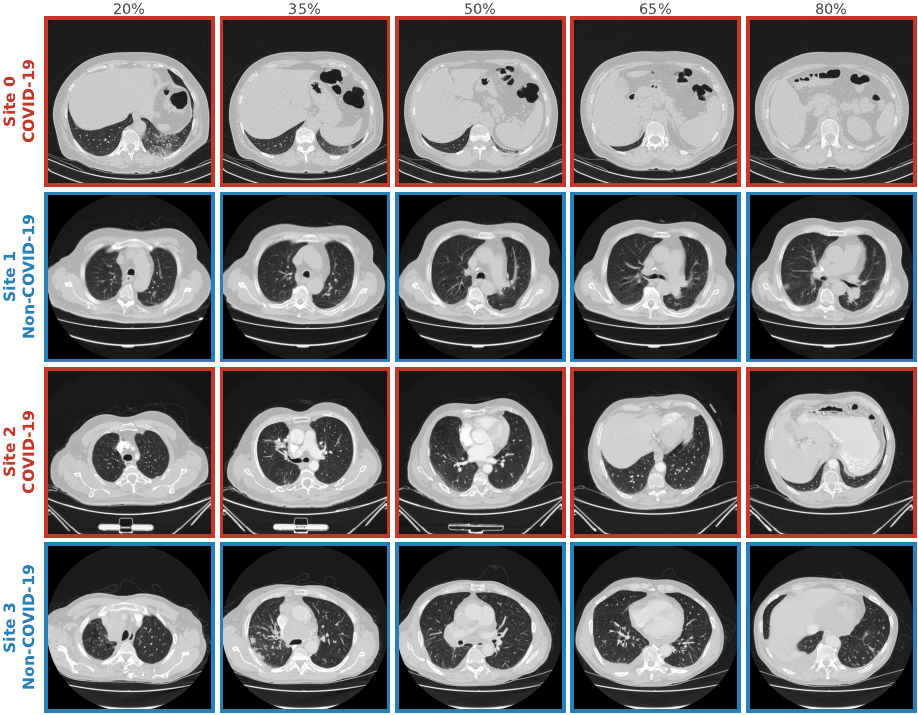}
\caption{Representative axial CT slices at five depth levels (20\%–80\%)
         for each acquisition site. Red borders indicate COVID-19 scans,
         blue borders Non-COVID-19, illustrating inter-site variation in
         image intensity and contrast.}
\label{fig:site_comparison}
\end{figure}

\begin{figure*}[t]
    \centering
    \includegraphics[width=\textwidth]{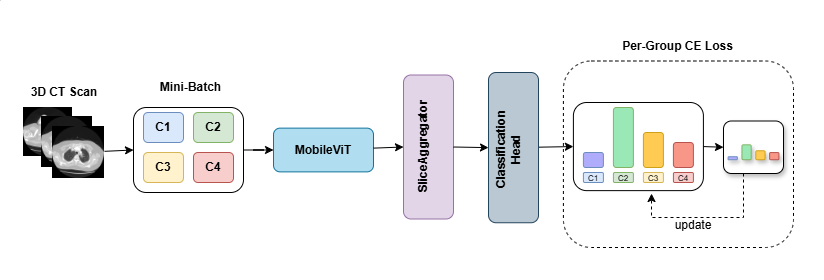}
    \caption{Overview of the proposed pipeline. A 3D CT scan is split into 64 slices, each encoded independently by a shared MobileViT backbone. A learned SliceAggregator pools slice features via attention weighting. The Classification Head produces predictions per sample, and Group DRO with KL regularization dynamically reweights per-centre losses during training.}
    \label{fig:architecture}
\end{figure*}
\section{Method}
\label{sec:method}

\subsection{Overview}

Our framework addresses volumetric CT classification under two forms of distribution shift: acquisition-site heterogeneity across four hospitals (\textit{Task~1}) and demographic imbalance across gender and class combinations (\textit{Task~2}). Figure~\ref{fig:site_comparison} shows representative CT slices illustrating inter-site variability for  \textit{Task~1}. The pipeline consists of three components: a slice-level visual encoder, a volumetric aggregator, and a domain-robust training objective.
Given a 3D CT volume, the model produces a class prediction of presence or absence of Covid-19 in a single forward pass, without test-time adaptation or site-specific normalisation.

\subsection{Problem Formulation}
 
Let $\mathcal{D} = \{(\mathbf{X}_i, y_i, g_i)\}_{i=1}^{N}$
denote a dataset of CT volumes, where
$\mathbf{X}_i \in \mathbb{R}^{1 \times S \times H \times W}$,
$y_i \in \mathcal{Y}$ is the class label, and
$g_i \in \mathcal{G}$ is a group identifier, where $S$ is the number of slices(64 in our case), H and W are the height and width of each slice, respectively.
For Task~1, $\mathcal{G} = \{0,1,2,3\}$ indexes the four
acquisition centres; for Task~2,
$\mathcal{G} = \{0,\ldots,7\}$ indexes the
$|\text{gender}| \times |\mathcal{Y}|= 2 \times 4$ subgroups.

Standard Empirical Risk Minimisation (ERM) minimises the
average loss $\frac{1}{N}\sum_i \ell(\mathbf{X}_i, y_i)$,
which can yield models that perform well in aggregate while
failing on minority groups~\cite{sagawa2020distributionally}.
We instead optimise for the worst-case group loss:
\begin{equation}
  \min_\theta \; \max_{g \in \mathcal{G}} \;
  \mathbb{E}_{(\mathbf{X},y)\sim\mathcal{D}_g}
  \bigl[\ell_\theta(\mathbf{X}, y)\bigr].
  \label{eq:dro_obj}
\end{equation}

In the task, our model learns to predict the presence or absence of COVID-19 given the CT scan of a subject.
 
\subsection{Model Architecture}
\label{sec:arch}
 The overall pipeline of the proposed method is illustrated in Figure~\ref{fig:architecture}; each component is described in detail in the following paragraphs.
\paragraph{Slice encoder.}
Each axial slice $\mathbf{x}_s \in \mathbb{R}^{1 \times H \times W}$ is processed independently by a shared MobileViT-XXS backbone~\cite{mehta2022mobilevit}. MobileViT-XXS combines lightweight MobileNet convolutions with
local Transformer blocks, yielding a strong feature extractor at only $\sim$1.3M parameters.
The original RGB stem is adapted to single-channel input by averaging the pretrained convolutional weights across the channel dimension, preserving pretrained representations. With the classification head removed, the backbone produces
a slice-level embedding
$\mathbf{f}_s = \text{Enc}(\mathbf{x}_s) \in \mathbb{R}^{d}$, where $d = 320$.
 
\paragraph{SliceTransformer Aggregator}
The $S = 64$ per-slice embeddings are stacked into a sequence
$\mathbf{F} = [\mathbf{f}_1, \ldots, \mathbf{f}_S]
\in \mathbb{R}^{S \times d}$
and passed through a two-layer Transformer Encoder with $d_{\text{ff}} = 4d$ and 4 attention heads. Global mean pooling over the sequence dimension yields a volume-level representation $\mathbf{z} \in \mathbb{R}^{d}$. Self-attention over the slice dimension allows the aggregator to learn which anatomical levels carry the strongest diagnostic signal, rather than treating all slices equally which could hurt the learning.

\paragraph{Classification head.}
The volume-level representation $\mathbf{z} \in \mathbb{R}^{d}$ is passed through a linear classifier preceded by LayerNorm and Dropout ($p\!=\!0.3$). For \textit{Task~1}, the head produces two logits (COVID / non-COVID); For \textit{Task~2}, it produces four logits corresponding to Adenocarcinoma, Squamous cell carcinoma, COVID-19, and Healthy. In both cases, class probabilities are obtained by applying softmax to the output logits of the linear head.
\subsection{Domain-Robust Training}
\label{sec:dro}
 
\paragraph{Group DRO with KL regularisation.}
We solve the minimax objective in Eq.~\eqref{eq:dro_obj}
via the exponentiated gradient update of~\cite{sagawa2020distributionally}.
Group weights $\{w_g\}$ are initialised to uniform
$w_g = 1/|\mathcal{G}|$ and updated each mini-batch:
\begin{equation}
  w_g \leftarrow
    \frac{w_g \cdot \exp(\eta_{\text{dro}} \cdot \ell_g)}
         {\sum_{g'} w_{g'} \cdot \exp(\eta_{\text{dro}} \cdot \ell_{g'})},
\end{equation}
where $\eta_{\text{dro}}\!=\!0.01$ is the DRO step size.
Groups absent from a mini-batch contribute $\ell_g\!=\!0$,
so their weights decay passively through renormalisation.
The group losses $\ell_g$ in the weight update are detached
from the computation graph; only $\mathcal{L}$ in
Eq.~\eqref{eq:loss} provides gradients to model parameters.
To prevent weight collapse onto a single hard group, we add a
KL penalty toward the uniform distribution:
\begin{equation}
  \mathcal{L} = \sum_{g \in \mathcal{G}} w_g \,\ell_g
    + \alpha \cdot D_{\mathrm{KL}}\!\left(\mathbf{w} \,\|\, \mathbf{u}\right),
  \label{eq:loss}
\end{equation}

where $\ell_g$ is the mean cross-entropy loss over
all samples in group $g$, $\mathbf{u}$ is the uniform distribution over groups and $\alpha \geq 0$ controls regularisation strength.
Setting $\alpha\!=\!0$ recovers standard \textit{vanilla} Group DRO;
large $\alpha$ forces equal group weighting, recovering empirical risk minimization(ERM).

\paragraph{Group definitions.}
For \textit{Task~1}, each group $g \in \{0,1,2,3\}$ corresponds to an
acquisition centre, directly matching the challenge evaluation
metric.
For \textit{Task~2}, each group $g = 2j + k$ is the joint index of
gender $j \in \{0,1\}$ (male/female) and class
$k \in \{0,1,2,3\}$ (Adenocarcinoma, Squamous, COVID-19,
Healthy), yielding $|\mathcal{G}|\!=\!8$ groups.
This fine-grained grouping explicitly upweights
underrepresented gender-class combinations such as
Female Squamous cell carcinoma ( in \textit{Task~2}),
which would otherwise be dominated by the loss on
majority classes.
 

\section{Experiments}
\label{sec:experiments}

\subsection{Datasets}

 \begin{table}[h]
\centering
\renewcommand{\arraystretch}{1.2}
\begin{tabular}{llrrr}
\toprule
\textbf{Centre} & \textbf{Class} & \textbf{Train} & \textbf{Val} & \textbf{Total} \\
\midrule
\multirow{2}{*}{0}
  & COVID-19     & 175 & 43 & 218 \\
  & Non-COVID-19 & 164 & 45 & 209 \\
\midrule
\multirow{2}{*}{1}
  & COVID-19     & 175 & 43 & 218 \\
  & Non-COVID-19 & 165 & 45 & 210 \\
\midrule
\multirow{2}{*}{2}
  & COVID-19     &  39 &  0 &  39 \\
  & Non-COVID-19 & 165 & 45 & 210 \\
\midrule
\multirow{2}{*}{3}
  & COVID-19     & 175 & 42 & 217 \\
  & Non-COVID-19 & 165 & 45 & 210 \\
\midrule
\multirow{2}{*}{\textbf{Total}}
  & \textbf{COVID-19}     & \textbf{564} & \textbf{128} & \textbf{692} \\
  & \textbf{Non-COVID-19} & \textbf{659} & \textbf{180} & \textbf{839}\\
\bottomrule
\end{tabular}
\caption{Dataset distribution of CT scans by data centre, class, and split.}
\label{tab:dataset_distribution}
\end{table}

\paragraph{Task 1: Multi-Source COVID-19 Detection}
This task includes CT volumes collected from four hospitals and medical
centres (centres 0-3) as shown in~\ref{tab:dataset_distribution}.
The dataset comprises 1{,}530 volumes
(train: 1{,}222; val: 308),
with 659 non-COVID and 564 COVID-positive cases in the
training split.
Each volume is labelled binary (COVID / non-COVID) and
annotated with a centre identifier $g \in \{0,1,2,3\}$.
The metric used for this task is the average macro F1 across all four
centres:
\begin{equation}
  P = \frac{1}{4}\sum_{i=0}^{3}
      \frac{F1^i_{\mathrm{covid}} + F1^i_{\mathrm{non\text{-}covid}}}{2}
\end{equation}
 \begin{figure}
    \centering
    \includegraphics[width=1.0\linewidth]{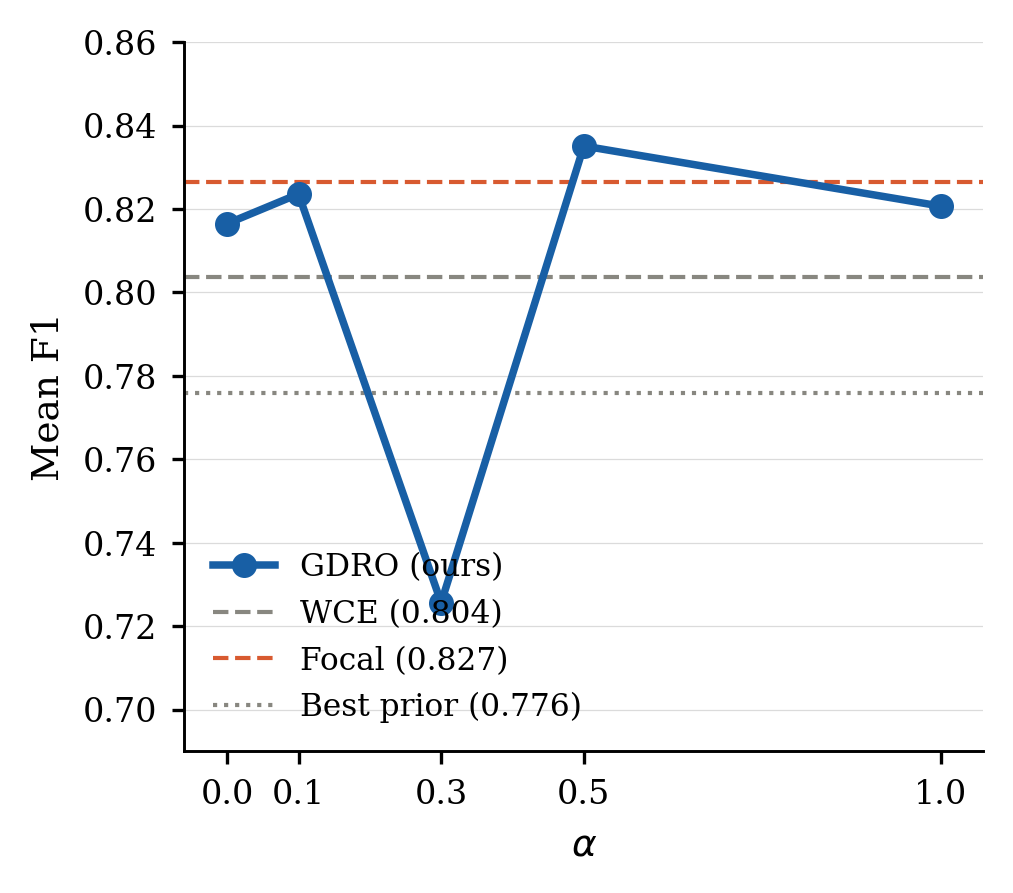}
    \caption{Effect of KL regularisation strength $\alpha$ on Task~2
    validation performance, reported separately for male and female
    subgroups. Group DRO with $\alpha\!=\!0.5$ achieves the best
    mean F1 of 0.815 and the smallest gender gap, outperforming
    Focal Loss (0.777) and the best challenge entry
    (0.704~\cite{kollias2025pharos}). At $\alpha\!=\!1.0$, male
    macro rises while female macro falls sharply, indicating that
    forcing uniform weights over-regularises the minority gender
    subgroup.}
    \label{fig:fig1_task2}
\end{figure}

\paragraph{Task 2: Fair Disease Diagnosis}
CT volumes are labelled across four classes: Adenocarcinoma (A), Squamous cell carcinoma (G), COVID-19, and Healthy (normal).
The dataset comprises 889 volumes (train: 734; val: 155) with notable class imbalance: female Squamous has only 5 training samples vs.\ 125 for female Adenocarcinoma. Each sample carries a gender label (male/female) in addition to the class label.
The evaluation metric is the mean per-gender macro F1.
\begin{equation}
  P = \frac{1}{2}(
      F1^{macro}_{\mathrm{male}} + F1^{macro}_{\mathrm{non\text{-}female}})
\end{equation}

\subsection{Implementation Details}

\begin{table}[t]
  \centering
  \small
  \setlength{\tabcolsep}{4pt}
  \renewcommand{\arraystretch}{1.2}

  \begin{tabular}{lccccc}
    \toprule
    \textbf{Method} &
    \textbf{C0} & \textbf{C1} & \textbf{C2} & \textbf{C3} &
    \textbf{$P$}$\uparrow$ \\
    \midrule
    \rowcolor{secblue}
    \multicolumn{6}{l}{\textit{\textcolor{blue!50!black}{Prior work}}} \\[1pt]
    \quad ACVLAB~\cite{lee2025multisource}
        & \underline{0.952} & 0.742 & 0.512 & 0.909 & 0.776 \\
    \quad FDVTS~\cite{yuan2025multisourcevre}
        & \textbf{0.962} & 0.729 & 0.490 & 0.921 & 0.776 \\
    \quad PHAROS~\cite{kollias2025pharos}
        &  0.858 & 0.653 & 0.442 & 0.851 &
        0.701\\
    \midrule
    \rowcolor{secgreen}
    \multicolumn{6}{l}{\textit{\textcolor{green!40!black}{Ours}}} \\[1pt]
    \quad WCE
        & 0.920 & 0.841 & \textbf{0.489}
        & \underline{0.965} & 0.804 \\
    \quad Focal ($\gamma\!=\!2$)
        & 0.920 & \underline{0.943} & \textbf{0.489}
        & 0.954 & \underline{0.827} \\
    \quad GDRO $\alpha\!=\!0.0$
        & 0.909 & 0.920 & \underline{0.483}
        & 0.954 & 0.817 \\
    \quad GDRO $\alpha\!=\!0.1$
        & 0.943 & 0.909 & 0.477
        & \underline{0.965} & 0.824 \\
    \rowcolor{bestrow}
    \quad GDRO $\alpha\!=\!0.5$
        & 0.920 & \textbf{0.955} & \textbf{0.489}
        & \textbf{0.977} & \textbf{0.835} \\
    \bottomrule
  \end{tabular}
  \vspace{3pt}
  \caption{Task~1 results and comparison with prior
           work~\cite{kollias2025pharos}.
           Per-centre macro F1 and overall $P\uparrow$.
           \textbf{Bold} = best; \underline{underline} = second best.}
  \label{tab:task1}
\end{table}

\begin{figure}
    \centering
    \includegraphics[width=1.0\linewidth]{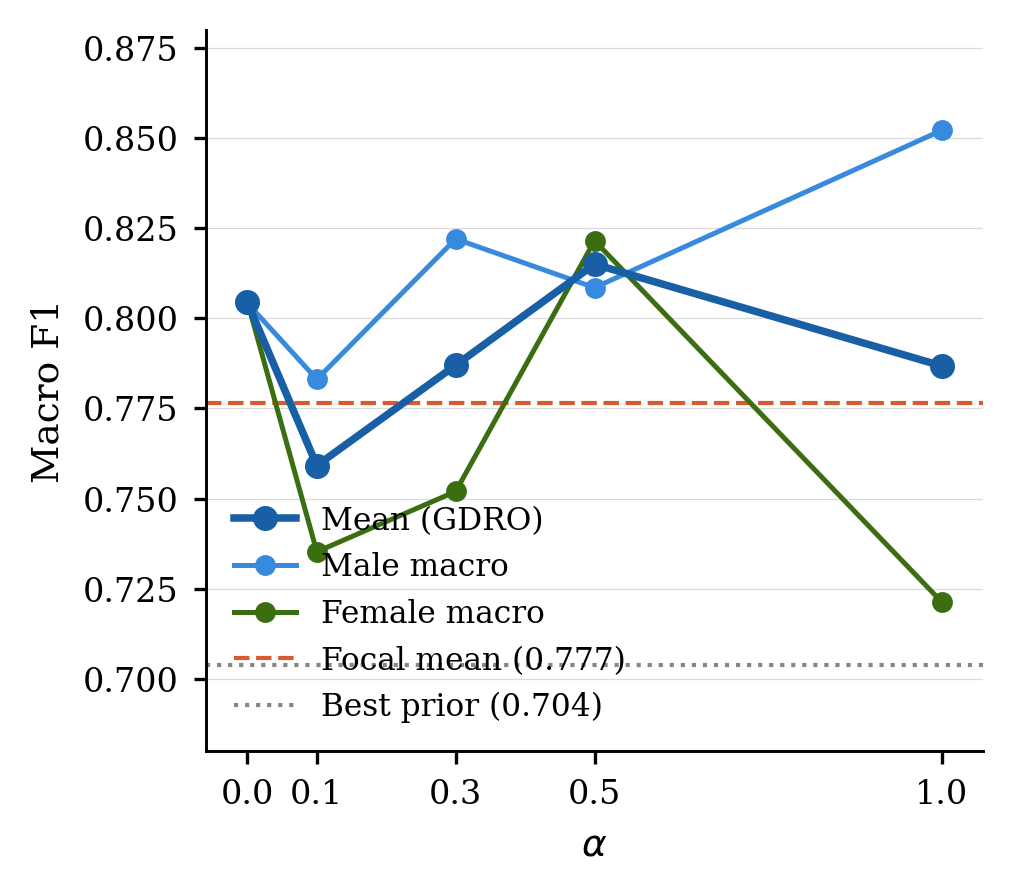}
    \caption{Effect of KL regularisation strength $\alpha$ on Task~1
validation performance. Group DRO with $\alpha\!=\!0.5$ achieves
the best mean F1 of 0.835, surpassing both the weighted CE
baseline (0.804) and the best published challenge entry
(0.776~\cite{kollias2025pharos}). Large $\alpha$ forces uniform
group weights, collapsing toward ERM and degrading performance
($\alpha\!=\!0.3$, F1\,=\,0.726).}
    \label{fig:fig1_task1}
\end{figure}

All models use MobileViT-XXS pretrained on ImageNet-1k. Input volumes are resampled to $S\!=\!64$ slices at $224 \times 224$ pixels. We used a batch size of 8 volumes; the SliceTransformer processes
$8 \times 64 = 512$ slice embeddings per step.
Training uses AdamW ($\beta_1\!=\!0.9$, $\beta_2\!=\!0.999$, weight decay $10^{-4}$) with a cosine schedule and a 5-step linear warm-up and an initial learning rate $10^{-4}$ unless stated otherwise. All experiments are conducted on a single NVIDIA L40S GPU. Early stopping monitors validation loss with a patience of 10 epochs.

\begin{table*}[t]
  \centering
  \small
  \setlength{\tabcolsep}{3.5pt}
  \renewcommand{\arraystretch}{1.2}
  \begin{tabular}{llccccc ccccc c}
    \toprule
    \multirow{2}{*}{\textbf{Configuration}} &
    \multirow{2}{*}{\textbf{Loss}} &
    \multicolumn{5}{c}{\cellcolor{secblue}\textbf{Male}} &
    \multicolumn{5}{c}{\cellcolor{secgreen}\textbf{Female}} &
    \multirow{2}{*}{\textbf{Mean F1}$\uparrow$} \\
    \cmidrule(lr){3-7}\cmidrule(lr){8-12}
    & &
    \textbf{A}$\uparrow$ & \textbf{G}$\uparrow$ & \textbf{Cv}$\uparrow$ &
    \textbf{No}$\uparrow$ & \textbf{Mac.}$\uparrow$ &
    \textbf{A}$\uparrow$ & \textbf{G}$\uparrow$ & \textbf{Cv}$\uparrow$ &
    \textbf{No}$\uparrow$ & \textbf{Mac.}$\uparrow$ & \\
    \midrule

    \rowcolor{secblue}
    \multicolumn{13}{l}{\textit{\textcolor{blue!50!black}{Prior work}}} \\[1pt]

    \quad FDVTS~\cite{li2025advancing}
        & -- & {--} & {--} & {--} & {--} & {62.5}
        & {--} & {--} & {--} & {--} & {78.3} & 70.4 \\
    \quad PHAROS ~\cite{kollias2025pharos}
        & -- & {--} & {--} & {--} & {--} & {55.9}
        & {--} & {--} & {--} & {--} & {68.7} & {62.3} \\

    \midrule
    \rowcolor{secgreen}
    \multicolumn{13}{l}{\textit{\textcolor{green!40!black}{Ours}}} \\[1pt]
    \quad Weighted CE & WCE
        & 0.8462 & 0.6364 & 0.8500          & \textbf{0.8500} & 0.7956
        & 0.8136 & 0.2500 & \textbf{0.9524} & \textbf{0.9231} & 0.7348
        & 0.7652 \\
    \quad Focal ($\gamma\!=\!2$) & FL
        & 0.7660 & 0.6429 & \textbf{0.9000} & 0.8718          & 0.7952
        & 0.7200 & 0.4615 & 0.9268          & \textbf{0.9231} & 0.7579
        & 0.7765 \\
    \midrule
    \multicolumn{13}{l}{\textit{Group DRO ($\alpha$ sweep)}} \\[2pt]
    \quad $\alpha\!=\!0.0$ & GDRO
        & 0.8000 & 0.6667 & 0.9000 & 0.8500 & 0.8042
        & 0.8421 & 0.5263 & 0.9268 & 0.9231 & 0.8046
        & 0.8044 \\
    \quad $\alpha\!=\!0.1$ & GDRO
        & 0.8235 & 0.6087 & 0.8500 & 0.8500 & 0.7831
        & 0.7667 & 0.3333 & 0.9524 & 0.8889 & 0.7353
        & 0.7592 \\
    \quad $\alpha\!=\!0.3$ & GDRO
        & \textbf{0.8889} & \textbf{0.7000} & 0.8571 & 0.8421 & \textbf{0.8220}
        & 0.8333 & 0.3750 & 0.9048 & 0.8947 & 0.7520
        & 0.7870 \\
    \rowcolor{bestrow}
    \quad $\alpha\!=\!0.5$ & GDRO
        & 0.8679 & 0.6667 & 0.8421 & 0.8571 & 0.8085
        & \textbf{0.8519} & \textbf{0.6364} & 0.9091 & 0.8889 & \textbf{0.8215}
        & \textbf{0.8150} \\
    \quad $\alpha\!=\!1.0$ & GDRO
        & 0.8936 & 0.8148 & 0.8500 & 0.8500 & 0.8521
        & 0.7719 & 0.3158 & 0.9091 & 0.8889 & 0.7214
        & 0.7868 \\
    \bottomrule
  \end{tabular}
  \caption{Task~2 results and comparison with prior work.
           Per-class F1 for each gender group
           (\textbf{A} = Adenocarcinoma, \textbf{G} = Squamous,
           \textbf{Cv} = COVID-19, \textbf{No} = Normal),
           per-gender macro F1, and mean across genders
           (challenge metric $\uparrow$).
           Prior work Mean F1 reported in \% from the challenge
           report~\cite{kollias2025pharos};
           ours in decimal.
           \textbf{Bold} = best per column.}
  \label{tab:task2_main}
\end{table*}

 
\paragraph{Quantitative comparison.}
Table~\ref{tab:task1} reports per-centre F1 and overall
challenge score $P$ across all configurations and prior work.
Group DRO with $\alpha\!=\!0.5$ achieves the best overall
F1 of \textbf{0.835}, outperforming weighted CE by
$+3.1$\, Focal Loss by $+0.8$\, and the best published
challenge entry (FDVTS, 0.776) by $+5.9$.
 
\paragraph{Effect of KL regularisation.}
Figure~\ref{fig:fig1_task1} shows the full $\alpha$ sweep.
Setting $\alpha\!=\!0.0$ (plain Group DRO) already improves
over both baselines, confirming the value of group-aware
training even without regularisation.
A small KL penalty ($\alpha\!=\!0.1$) provides a further
gain, while $\alpha\!=\!0.3$ degrades sharply to 0.726
consistent with the KL term over-constraining group weight
dynamics and collapsing toward ERM behaviour.
The optimal $\alpha\!=\!0.5$ sits at the balance point
between worst-case protection and average performance
stability, a pattern that holds on Task~2 as well,
suggesting it is a robust hyperparameter choice for
small multi-site medical datasets.
 
\paragraph{Centre 2 domain shift.}
Centre~2 exhibits near-zero COVID F1 across \emph{all}
configurations (Table~\ref{tab:task1}).
Every method produces near-zero COVID F1 for Centre~2 while
centres 0, 1, and 3 each exceed 0.84.
This persistent failure is consistent with a severe
single-centre acquisition shift is likely a distinct scanner
model, reconstruction kernel, or imaging protocol, that
produces a distribution not represented in any other
training site.
Group DRO addresses imbalanced loss across known groups,
but it cannot recover signal from a distribution that was never observed during training. This result highlights a fundamental limitation of
reweighting-based approaches that we intend to address
in future work via test-time adaptation.

\subsection{Task 2: Fair Disease Diagnosis}
 
\paragraph{Quantitative comparison.}
Table~\ref{tab:task2_main} reports per-class and per-gender
F1 across all configurations.
Group DRO with $\alpha\!=\!0.5$ achieves the best mean
per-gender macro F1 of \textbf{0.815}, outperforming Focal
Loss and the weighted CE baseline, and surpassing the best challenge entry
(FDVTS, 70.4\%). Gains are consistent across both gender groups: male macro F1 improves from 0.795 (Focal) to 0.809,
and female macro from 0.758 to 0.822.
 
\paragraph{Effect of KL regularisation.}
Figure~\ref{fig:fig1_task2} shows the full $\alpha$ sweep,
reported separately for male and female subgroups.
Without regularisation ($\alpha\!=\!0.0$), Group DRO already
outperforms both baselines.
Increasing $\alpha$ to 0.5 yields the best mean F1 and the
smallest gender gap ($\Delta\!=\!0.013$), suggesting the
KL penalty suppresses weight collapse onto the hardest group
without sacrificing minority performance.
At $\alpha\!=\!1.0$, male macro rises to 0.852 while female
macro falls to 0.721, an artefact of forcing weights toward
uniformity, which disproportionately benefits majority
subgroups and reduces the targeted upweighting of
Female~Squamous.
 
\paragraph{Fine-grained group formulation.}
The 8-group (gender$\times$class) formulation is central
to the Task~2 result.
A coarser 2-group (gender-only) design would conflate
the female-Squamous subgroup with the much larger
female-Adenocarcinoma group (125 samples), preventing
targeted upweighting of the most underrepresented
combination.
The improvement in Female-G F1 from 0.462 (Focal Loss)
to 0.636 (GDRO $\alpha\!=\!0.5$) is the clearest evidence that fine-grained
grouping, not simply better average optimisation, drives
the performance gain on the minority subgroup.

\section{Conclusion}
\label{sec:conclusion}
 
We presented a domain-robust and fairness-aware framework
for volumetric CT classification, combining a lightweight
MobileViT-XXS slice encoder with a SliceTransformer
aggregator and a KL-regularised Group DRO training
objective.
On Task~1 (COVID-19, four acquisition centres), our best
configuration achieves a challenge F1 of \textbf{0.835},
outperforming the best published challenge entry by
$+5.9$\,pp.
On Task~2 (four-class lung pathology with gender fairness
constraints), Group DRO with $\alpha\!=\!0.5$ achieves a
mean per-gender macro F1 of \textbf{0.815}, surpassing the
best challenge entry  and improving Female
Squamous F1 over the Focal Loss baseline.
 
The persistent failure on Centre~2 in \textit{Task~1} highlights
that reweighting-based approaches cannot compensate for
domains that are absent from training data.
Future work will explore test-time adaptation and
site-specific normalisation as complementary strategies
for this failure mode. Looking ahead, a natural direction for this work is model
compression toward deployment on lower-powered edge devices, such as point-of-care scanners and portable imaging units common in resource-constrained clinical settings.
{
    \small
    \bibliographystyle{ieeenat_fullname}
    \bibliography{main}
}


\end{document}